\documentclass[11pt]{article} 
\usepackage{rldmsubmit,palatino}
\usepackage{amsmath}
\usepackage{amssymb}
\usepackage{algorithm}
\usepackage[noend]{algpseudocode}
\usepackage{ dsfont }
\usepackage{subfig}
\usepackage{graphicx}
\usepackage[backend=biber]{biblatex}
\usepackage{wrapfig}

\title{Predecessor Features}

\author{
Duncan Bailey\\
Department of Cognitive Science\\
University of California\\
San Diego\\
\texttt{dbailey@ucsd.edu}
\And
Marcelo G. Mattar\\
Department of Cognitive Science\\
University of California\\
San Diego\\
\texttt{mmattar@ucsd.edu} \\
}

\begin{document}

\maketitle

\begin{abstract}

Any reinforcement learning system must be able to identify which past events contributed to observed outcomes, a problem known as credit assignment. A common solution to this problem is to use an eligibility trace to assign credit to recency-weighted set of experienced events. However, in many realistic tasks, the set of recently experienced events are only one of the many possible action events that could have preceded the current outcome. This suggests that reinforcement learning can be made more efficient by allowing credit assignment to any viable preceding state, rather than only those most recently experienced. Accordingly, we examine ``Predecessor Features'', the fully bootstrapped version of van Hasselt's ``Expected Trace'', an algorithm that achieves this richer form of credit assignment. By maintaining a representation that approximates the expected sum of past occupancies, this algorithm allows temporal difference (TD) errors to be propagated accurately to a larger number of predecessor states than conventional methods, greatly improving learning speed. The algorithm can also be naturally extended from tabular state representation to feature representations allowing for increased performance on a wide range of environments. We demonstrate several use cases for Predecessor Features and compare its performance with other approaches.


\end{abstract}

\keywords{
Reinforcement Learning, TD-Learning, Eligibility Traces, Successor Representation
}

\acknowledgements{We would like to thank the Cognitive Science Department of the University of California San Diego.}

\startmain 
\section{Introduction}

At the core of Reinforcement Learning (RL) is the problem of credit assignment: identifying which of the previously chosen events (actions and states) are causally related to an observed outcome. In the temporal difference (TD) learning algorithm, a cornerstone of RL, recently experienced events receive credit for the prediction errors encountered at each present moment. Thus, to distribute credit appropriately, a temporary record of recently experienced events -- an eligibility trace -- is maintained and updated continuously. 

The assumption underlying algorithms based on eligibility traces is that only recent events can receive credit for an outcome. However, in many realistic tasks, the set of recently experienced events are only one of the many possible action events that could have preceded the current outcome. This suggests that reinforcement learning can be made more efficient by allowing credit to be assigned to any viable preceding state, rather than only those most recently experienced. But how to identify the correct set of preceding states in this richer form of credit assignment? To help answer this question we consider first the Successor Representation (SR), which quantifies, from each state, the expected (discounted) future occupancy of every other state \cite{dayan1993}. This representation can be leveraged to compute values from rewards through a simple linear operation, and can be learned efficiently in the setting of linear function approximation \cite{barreto2019transfer}. We propose that this representation can also express, from each state, the set of all possible (discounted) preceding states, yielding a quantity analogous to an eligibility trace.

Here, we formalize precisely the relationship between the SR and the (expectation of) eligibility traces. We then examine a variant of temporal difference learning that uses this richer form of eligibility traces, an algorithm we call Predecessor Representation. To extend this method to the setting of function approximation, we describe an approach to learn the predecessors directly, giving rise to a second algorithm called Predecessor Features. This ends up being a special case of the Expected traces algorithm \cite{et}. In both cases, we demonstrate a few simulations showing that this learned quantity is helpful for credit assignment as it acts as an improved way of propagating error to relevant features than standard approaches. 

\section{Formalism}

The central task in Reinforcement Learning is to predict returns of future discounted rewards:
$
G_{t:T} =\sum_{i=1}^{T} \gamma_{t+i}^{(i-1)} R_{t+i}
$
where $T$ is the time the current episode terminates or $T=\infty$ for continuous tasks. The value $v_{\pi}(s)=\mathbb{E}_{\pi}\left[G_{t:T} \mid S_{t}=s\right]$ of state $s$ is the expected return for a policy $\pi$ when starting from state $s$. This value is approximated by the function $v_{\mathrm{w}}(s) \approx v_{\pi}(s)$, parameterized by a weight vector $\mathbf{w} \in \mathbb{R}^{d}$. The value function can be specified as a table with each entry corresponding to a state, as a linear function of some defined input features, or as a non-linear function. The weight vector $\mathbf{w}$ is learned iteratively through the update rule
$
\mathbf{w}_{t+1}=\mathbf{w}_{t}+\Delta \mathbf{w}_{t}
$
such that $v_{\mathbf{w}}$ approaches the true $v_{\pi}$. 

There are multiple methods for computing $\Delta \mathbf{w}_{t}$. The Monte Carlo algorithm defines $\Delta \mathbf{w}_{t} \equiv \alpha\left(R_{t+1}+\gamma_{t+1} G_{t+1: T}-v_{\mathbf{w}}\left(S_{t}\right)\right) \nabla_{\mathbf{w}} v_{\mathbf{w}}\left(S_{t}\right)$ and is able to succeed in the task of value function approximation, but does so with high variance. An alternative approach that estimates the value function with lower variance is TD learning. TD learning `learns a guess from a guess' and replaces the return with the current expectation $v\left(S_{t+1}\right) \approx G_{t+1: T}$, where $\Delta \mathbf{w}_{t} \equiv \alpha\left(R_{t+1}+\gamma_{t+1} v_{\mathbf{w}}\left(S_{t+1}\right)-v_{\mathbf{w}}\left(S_{t}\right)\right)\nabla_{\mathbf{w}} v_{\mathbf{w}}\left(S_{t}\right)$.

Algorithms that employ a `forward' looking approach, like the MC algorithm, use returns that depend on future trajectories and need to wait many time steps or until the end of an episode to create their updates. Alternatively, algorithms that employ a `backward view' will look back on past experiences during an episode to update current estimates. For example, a classic backward-looking algorithm, $\operatorname{TD}(\lambda)$, defines
$
 \quad \Delta \mathbf{w}_{t} \equiv \alpha \delta_{t} \boldsymbol{e}_{t} \quad
$
where $\boldsymbol{e}_{t}=\gamma_{t} \lambda \boldsymbol{e}_{t-1}+\nabla_{\mathbf{w}} v_{\mathbf{w}}\left(S_{t}\right)$ is referred to as eligibility trace, and $\delta_{t} = R_{t+1}+\gamma_{t+1} v_{\mathbf{w}}\left(S_{t+1}\right)-v_{\mathbf{w}}\left(S_{t}\right)$ is referred to as the temporal difference (TD) error.

$\operatorname{TD}(\lambda)$ generalizes Monte Carlo and Temporal Difference methods, with $\operatorname{TD}(1)$ corresponding to an online implementation of the MC algorithm and $\operatorname{TD}(0)$ corresponding to a regular one-step TD update. 

\section{Predecessor Representation}

We first offer a generalization of the concept of eligibility traces in the tabular case, leading to an algorithm called ``Temporal differences -- Predecessor Representation'', or TD-PR. In the original $\operatorname{TD}(\lambda)$ algorithm, the main way of propagating value updates to relevant states is by keeping an eligibility trace. In the tabular $\operatorname{TD}(\lambda)$, the eligibility trace is defined by $e_{t}(s)= \begin{cases}\gamma \lambda e_{t-1}(s) & \text { if } s \neq s_{t} \\ \gamma \lambda e_{t-1}(s)+1 & \text { if } s=s_{t}\end{cases}$ where on each step the eligibility traces for all states decay by $\gamma\lambda$, and the eligibility trace for the one state visited on the step is incremented by 1. The eligibility trace defines the extent to which each state is `eligble' for undergoing learning changes. The corresponding TD-error is $\delta_{t}=R_{t+1}+\gamma V_{t}\left(S_{t+1}\right)-V_{t}\left(S_{t}\right)$ and value function update $\Delta V_{t}(s)=\alpha \delta_{t} e_{t}(s), \text { for all } s \in \mathcal{S}$.

Our starting point to generalize the concept of eligibility traces is the Successor Representation (SR) \cite{dayan1993}. Given a stream of experience, the SR maitrx $\mathbf{M}$ represents a given state in terms of discounted occupancy to other states. 
The SR is defined as $\mathbf{M}=(\mathbf{I}-\gamma \mathbf{P})^{-1}$, where $\mathbf{P}$ is a transition matrix with entries corresponding to probabilities of transitioning from states (or state action pairs) to other states. 
Thus, if $\mathbf{M}_{i j} $ represents the expected (discounted) number of visitations from i to j, the ith row of $\mathbf{M}$ represents the expected (discounted) number of visitations from i to every state. Accordingly, the jth column of $\mathbf{M}$ represents the expected (discounted) number of visitations to state j, starting from every state. Indeed, while a row of $\mathbf{M}$ is ``forward-looking'', a column of $\mathbf{M}$ is ``backward-looking''. If we set the discount factor to $\gamma\lambda$:
\begin{align*}
    \mathbf{M}_{i j}&=\mathds{E} \left[\sum_{p=0}^{\infty} (\gamma\lambda)^{p} \mathds{1}_{s_{\{n+p+1} = j\}} \mid s_{n}=i\right]
=\sum_{p=0}^{\infty} (\gamma\lambda)^{p} \mathds{P}\left(s_{n+p+1}=j \mid s_{n}=i\right) =\frac{\mathds{P}(s=j)}{\mathds{P}(s=i)} \mathds{E}\left[\sum_{p=0}^{\infty} (\gamma\lambda)^{p} \mathds{1}_{\{s_{n-p-1} = i\}} \mid s_{n}=j\right]\\
&=\frac{\mathds{P}(s=j)}{\mathds{P}(s=i)}\sum_{p=0}^{\infty}(\gamma\lambda)^{p} \mathds{P}\left(s_{n-p-1}=i \mid s_{n}=j\right)
=\frac{\mathds{P}(s=j)}{\mathds{P}(s=i)} \mathds{E}\left[\sum_{p=0}^{\infty} (\gamma\lambda)^{p} \mathds{1}_{\{s_{n-p-1} = i\}} \mid s_{n}=j\right]
=\frac{\mathds{P}(s=j)}{\mathds{P}(s=i)}
\mathds{E}\left[ e_{n-1}(i) \mid s_{n}=j\right] 
\end{align*}
In other words, the column of the SR is directly related to the expectation of the eligibility traces (see also van Hasselt et al \cite{et}, Pitis et al \cite{silviu}). In contrast to a sample of the eligibility traces, the expected trace has the advantage that is contains all possible predecessor states, and thus can give rise to a more efficient TD algorithm. 

Below we describe TD-PR, an example of such algorithm. 

\begin{algorithm}
\caption{Tabular TD-PR}\label{alg:cap}
\begin{algorithmic}[1]
\Procedure{TD-PR}{$episodes,\gamma,\lambda,\alpha,\beta$}
\State $\mathbf{v} \gets \mathbf{0}$
\State $\mathbf{M} \gets \mathbf{0}$ (\(|S| \times |S|\) identity matrix)
\For{episode in $1\ldots n$}
    (Note: $\mathbf{s}_{k}$ is one-hot vector with 1 at index k)
    \State $\mathbf{e} \gets \mathbf{0}$ (eligibilty trace)
    \For{pair $(s_{i}, s_{j} )$ and reward $r$ in episode}
    \State $e(s_{i}) \gets e(s_{i}) + 1$
    \State $\mathbf{M} \gets \mathbf{M}+\beta \mathbf{e} \mathbf{s}_{j}^{\intercal}+\beta \gamma \mathbf{e} \mathbf{s}_{j}^{\intercal} \mathbf{M}-\beta \mathbf{e} \mathbf{s}_{i}^{\intercal} \mathbf{M}$
    \State $\mathbf{v} \gets \mathbf{v}+\alpha\mathbf{M}_{\cdot i}(r + \gamma v_{j} - v_{i})$
    \State $\mathbf{e} \gets \gamma\lambda\mathbf{e}$
    \EndFor
\EndFor\\
\Return $\mathbf{v}, \mathbf{M}$
\EndProcedure
\end{algorithmic}
\end{algorithm}

Note that TD-PR learns the SR using temporal differences. At the end of the first episode, a column of the SR (predecessor representation) corresponds exactly to the usual eligibility trace. At the second episode and beyond, however, the two representations diverge. This is seen clearly in Figures 1b and 1c.

\begin{figure}[htp]
    \centering
    \subfloat[]{{\includegraphics[width=0.5\textwidth]{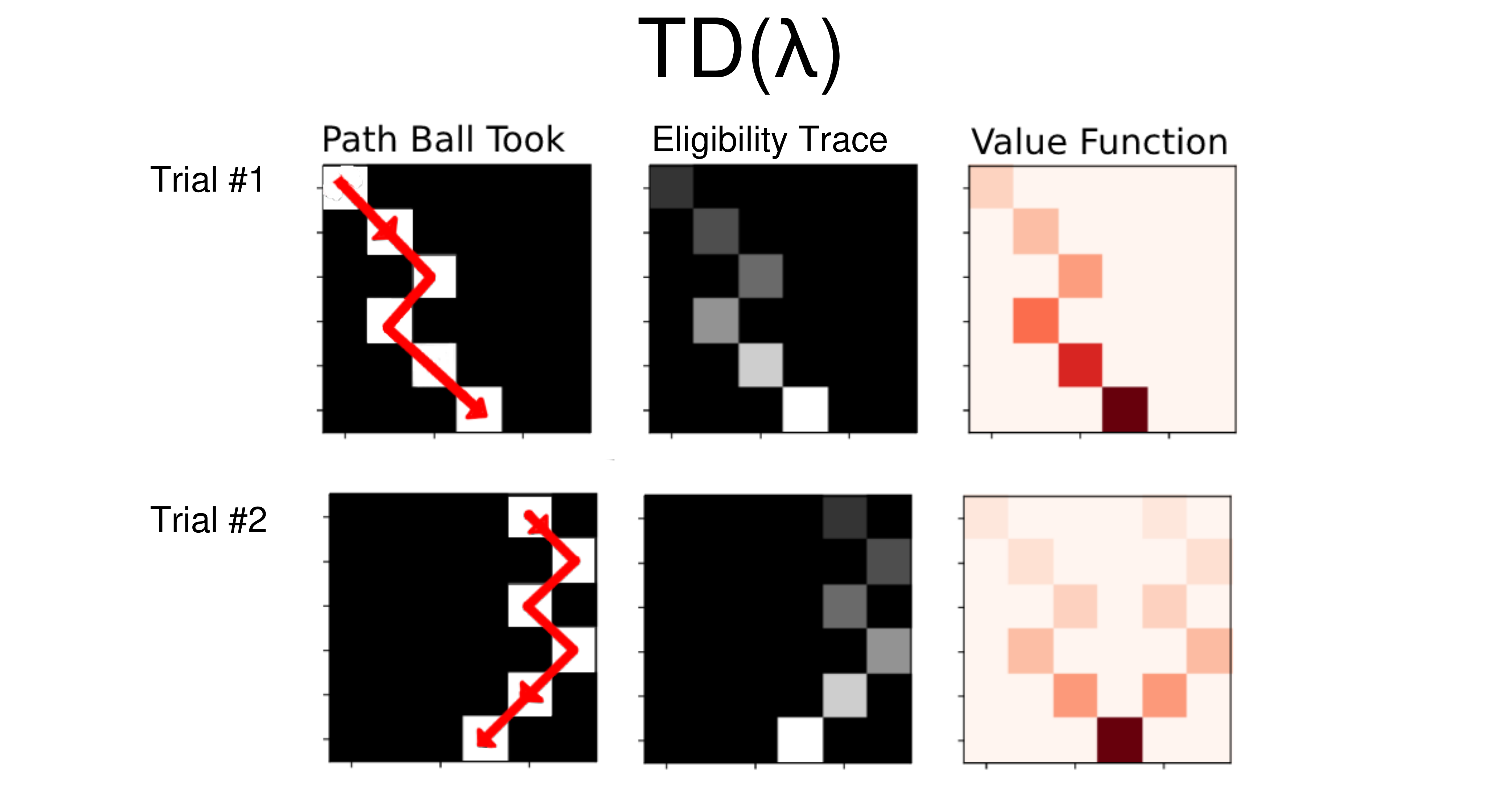} }}%
    \subfloat[]{{\includegraphics[width=0.5\textwidth]{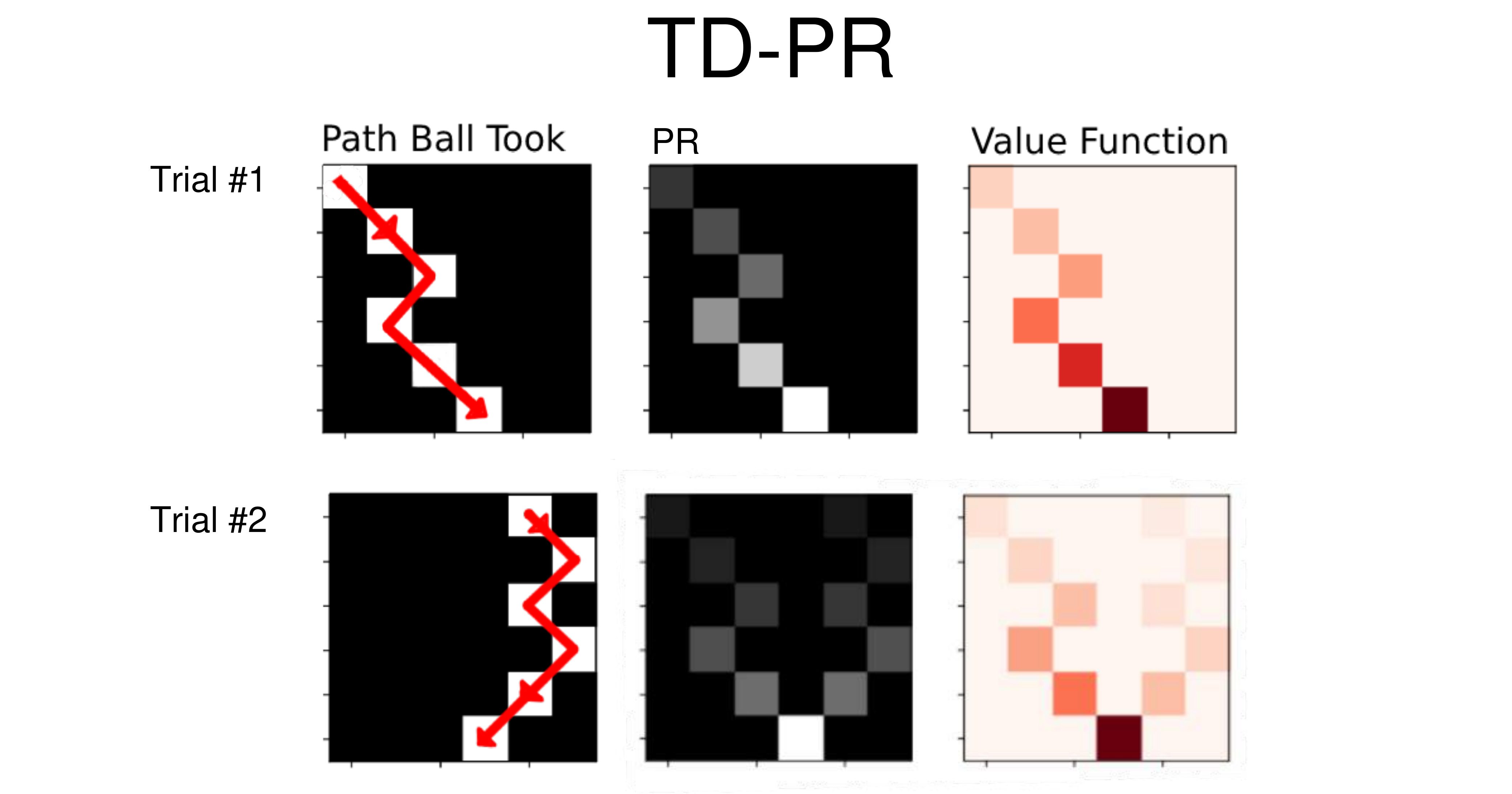} }}%
    \caption{Illustrated is the first two trials where the ball hits the reward. Displayed from left to right is the path the ball took, the credit assignment vector (shown as a  $6 \times 6$ matrix) used in each algorithm, and the value function at the end of the trial. It becomes apparent that the key difference between the two algorithms is TD-PR uses a credit assignment method that keeps a memory of past trial's state visits and updates all state values accordingly.}%
    \label{fig:example}%
\end{figure}

\section{Predecessor Features}

We now examine a generalization of the concept of eligibility traces in the general, function approximation case, leading to an algorithm called ``Temporal differences -- Predecessor Features'', or TD-PF. An expected eligibility trace must be defined as: $\boldsymbol{z}(s) \equiv \mathbb{E}\left[\boldsymbol{e}_{t} \mid S_{t}=s\right]$

This expected eligibility trace tries to approximate the sum of discounted predecessor features:
\begin{align*}
    \boldsymbol{z}(s)=\mathbb{E}\left[\sum_{n=0}^{\infty} \left(\lambda \gamma\right)^{n} \mathbf{x}(S_{t-n}) \mid S_{t}=s\right]
\end{align*}
 $\lambda \neq 1$ should be included if a partial predecessor feature representation is desired. In TD-PF we try to learn approximations $\boldsymbol{z}_{\boldsymbol{\theta}}\left(S_{t}\right) \approx \boldsymbol{z}\left(S_{t}\right)$ with parameters $\boldsymbol{\theta} \in \mathbb{R}^{d}$. In doing so we will be using supervised learning techniques to minimize the empirical loss $\mathcal{L}\left(\boldsymbol{y}, \boldsymbol{z}_{\boldsymbol{\theta}}\left(S_{t}\right)\right)$. Where $\boldsymbol{y}$ is the target for each collected transition $\left(s, a, r, s^{\prime}\right)$.
\begin{align*}
    \boldsymbol{y}= \begin{cases}\nabla_{\mathbf{w}} v_{\mathbf{w}}\left(S_{t}\right) & \text { if } S_{t} \text { is an initial state } \\ \nabla_{\mathbf{w}} v_{\mathbf{w}}\left(S_{t}\right)+\lambda \gamma  \boldsymbol{z}_{\boldsymbol{\theta}}\left(S_{t-1}\right) & \text { otherwise }\end{cases}
\end{align*}
Here $\mathbf{x}(S_{t})$ is a feature vector of $S_{t}$. In this case, we will be using a TD like update of $\boldsymbol{z}_{\boldsymbol{\theta}}\left(S_{t}\right)$. The learned representation will be map from a current feature vector to a vector of predecessor features. If we use a squared loss function, then
\begin{align*}
    \mathcal{L}\left(\boldsymbol{y}, \boldsymbol{z}_{\boldsymbol{\theta}}\left(S_{t}\right)\right)=\mathbb{E}\left[\left\|\boldsymbol{y}-\boldsymbol{z}_{\boldsymbol{\theta}}\left(S_{t}\right)\right\|^{2}\right]
    \\
    \Delta \boldsymbol{\theta} \leftarrow \nabla_{\boldsymbol{\theta}}\left\|\nabla_{\mathbf{w}} v_{\mathbf{w}}\left(S_{t}\right)+\lambda \gamma\boldsymbol{z}_{\boldsymbol{\theta}}(S_{t-1})-\boldsymbol{z}_{\boldsymbol{\theta}}(S_{t})\right\|_{2}^{2}
\end{align*}
For simplicity we will use a linear approximation of $\boldsymbol{z}_{\boldsymbol{\theta}}$ such that
$
    \boldsymbol{z}_{\boldsymbol{\theta}}\left(S_{t}\right)=\boldsymbol{\Psi} \mathbf{x}(S_{t})
$
where $\boldsymbol{\Psi}$ is a square matrix. Note that any approximation method could be used here (i.e. a neural network). Computing the gradient of the squared loss function with respect to $\boldsymbol{\Psi}$ and performing gradient descent we get an update rule similar to linear TD-learning:
\begin{align*}
    \boldsymbol{\Psi}_{t+1}=\boldsymbol{\Psi}_{t}-\beta\left(\boldsymbol{z}_{\boldsymbol{\theta}}\left(S_{t}\right)-\boldsymbol{y}_{s, a, r, s^{\prime}}\right) \mathbf{x}(S_{t})^{\intercal}
\end{align*}
Note that Lehnert and Littman showed a similar learning rule for an approximation of Successor Features \cite{lehnert2019successor}. This TD-inspired update of the expected trace parameters is a special instance of van Hasselt et al\cite{et} ET($\lambda, \eta$) algorithm where $\eta=0$ (full bootstrapping). This contrasts updating towards a sampled eligibility trace, which would reduce bias but increase variance of the approximation. It should be noted that simulations show that the $\mathbf{\Psi}$ matrix when using a tabular feature vector for each state approximately learns the SR matrix and gives similar performance to TD-PR.
\begin{algorithm}
\caption{Linear TD-PF}\label{alg:cap}
\begin{algorithmic}[1]
\Procedure{Linear TD-PF}{$episodes,\mathbf{w},\boldsymbol{\Psi},\alpha, \beta$}
\State initialize $\mathbf{w},\boldsymbol{\Psi}$
\For{episode in $1\ldots n$}
    \State $S_{t} \gets \text{initial state of episode}$
   \State $\mathbf{y} \gets \mathbf{x}(S_{t})$
   \State $\boldsymbol{\Psi}_{t+1}=\boldsymbol{\Psi}_{t}-\beta\left(\boldsymbol{z}_{\boldsymbol{\theta}}\left(S_{t}\right)-\mathbf{y}\right) \mathbf{x}\left(S_{t}\right)^{\boldsymbol{\top}}$ (Note: $\boldsymbol{z}_{\boldsymbol{\theta}}\left(S_{t}\right)=\boldsymbol{\Psi} \mathbf{x}(S_{t})$ )
    \For{pair $(S_{t}, S_{t+1} )$ and reward $r$ in episode}

    \State $\delta \gets r + \gamma v_{\mathbf{w}}\left(S_{t+1}\right)-v_{\mathbf{w}}(S_{t})$
    \State $\mathbf{y} \gets \mathbf{x}(S_{t+1}) + \lambda \gamma \boldsymbol{z}_{\boldsymbol{\theta}}\left(S_{t}\right)$
    \State $\boldsymbol{\Psi}_{t+1}=\boldsymbol{\Psi}_{t}-\beta\left(\boldsymbol{z}_{\boldsymbol{\theta}}\left(S_{t+1}\right)-\mathbf{y}\right) \mathbf{x}\left(S_{t+1}\right)^{\boldsymbol{\top}}$
    \State $\mathbf{w} \leftarrow \mathbf{w}+\alpha \delta \boldsymbol{z}_{\boldsymbol{\theta}}(S_{t})$
    \EndFor
\EndFor\\
\Return $\mathbf{w}, \boldsymbol{\Psi}$
\EndProcedure
\end{algorithmic}
\end{algorithm}

\section{Experiments}

\begin{figure}[htp]
    \centering
    \subfloat[]{{\includegraphics[width=0.3\textwidth]{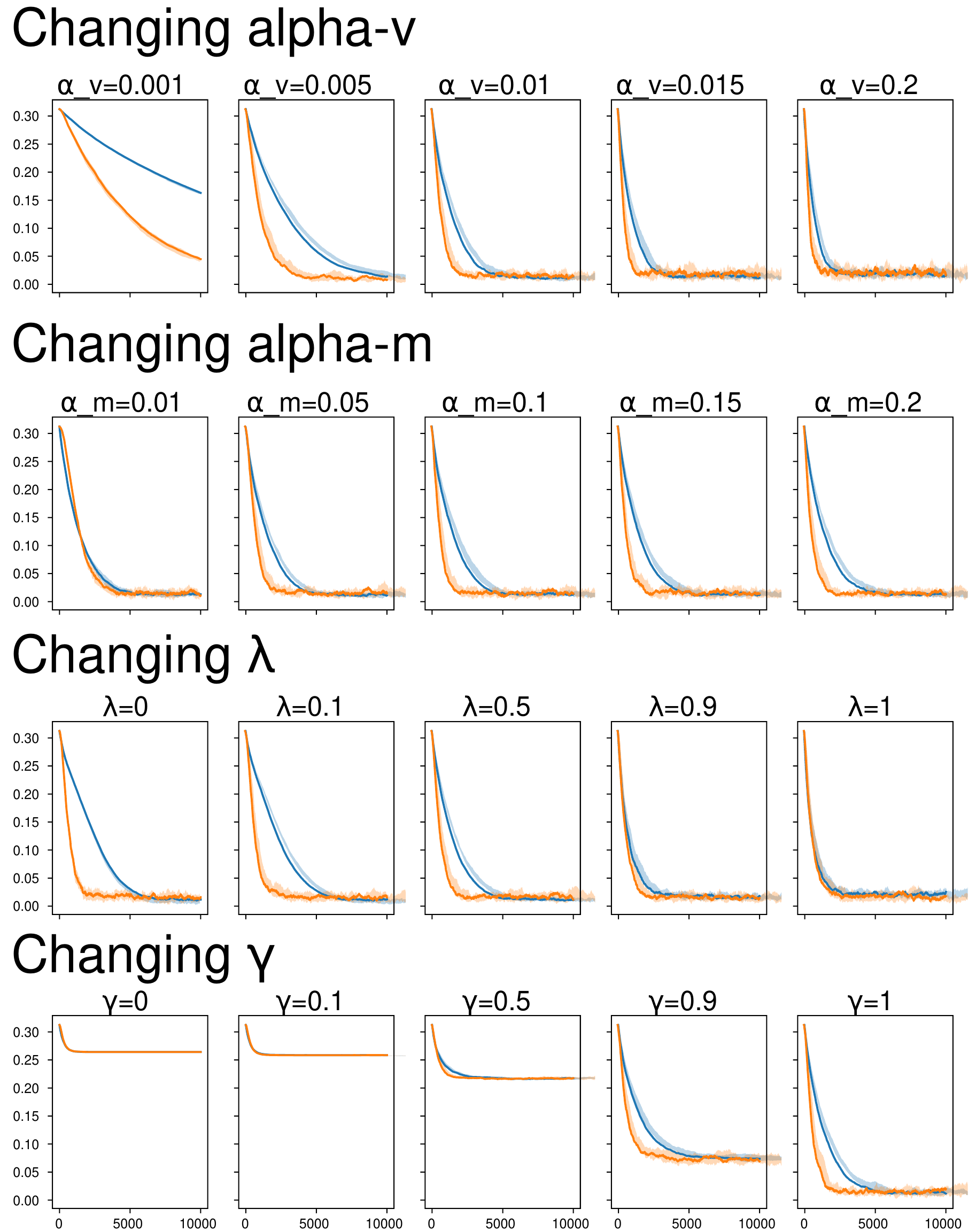} }}%
    \subfloat[]{{\includegraphics[width=0.4\textwidth]{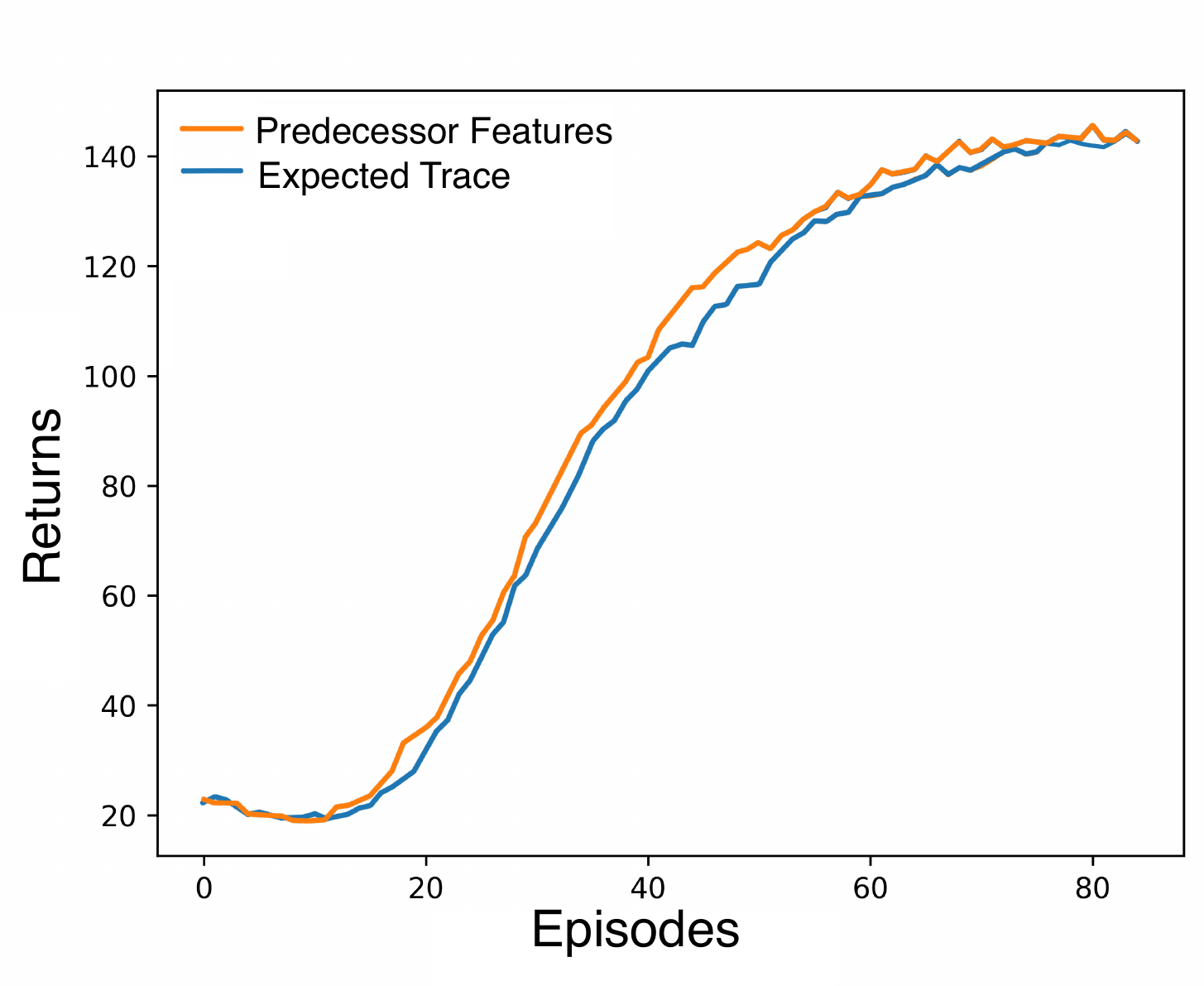} }}%
    \caption{
    \textbf{(a)} Comparison of convergence to true value function for TD-PR (in orange) and TD($\lambda$) (in blue). Each row refers to changing a different parameter of each function. Alpha-v and Alpha-m are the learning rates for the value function and SR matrix, respectively. Aside from the changing variable all other parameters were set to equivalent default values of (alpha-v = 0.01, alpha-m=0.1, $\lambda = 0.9$, $\gamma = 1$).
    \textbf{(b)} The ``Predecessor Features'' algorithm (a fully bootstrapped method) marginally outperforms the ``ExpectedTrace'' algorithm (no bootstrapping, ET($\lambda, \eta = 0$), the Monte Carlo update) using a deep neural network approximation for the value function. There is only a slight difference between the two algorithms, however it leads to a difference in performance. The y axis is returns on each episode and the x-axis is the number of episodes the agent has experienced.
    }%
    \label{fig:example}
\end{figure}
The task we studied the methods of TD-PR on is the Japanese gambling game Pachinko (aka Plinko). Plinko is a $6 \times 6$ grid where a ball is placed with $1/6$ probability into one of the top slots. A ball will go down a row and to the left with probability $1/2$ and down a row and to the right with probability $1/2$, unless the ball is at a state at the edge of the grid. In this case, the ball will go down a row and away from the edge by one state with probability $1$ on each time step. A reward of $0$ was received on each time step for every state unless the ball landed in the fourth state from the left on the bottom row where a reward of $1$ was received. Once the ball reached the final row, the episode was terminated. Every episode lasted 6 time steps and restarted from the top row. The state space had $36$ elements ($|S| = 36$), meaning the size of the SR matrix was $36 \times 36$. A column of the SR matrix was a $36 \times 1$ sized vector and could be nicely reformatted to fit the size of the Plinko board of size $6 \times 6$. 

In TD-PR applied to Plinko for value function approximation, after each step of each episode, the sample SR matrix is updated and a column of the sample SR is used to propagate the TD-error to qualifying states. In convergence to the true value function Figure 2 shows TD-PR outperforms $\operatorname{TD}(\lambda)$ for all parameter values. Note that the only parameter value where $\operatorname{TD}(\lambda)$ initially outperforms TD-PR is when alpha-m, representing the learning rate for the sample SR, is quite small (0.01), however it does converge to the optimal value function ahead of $\operatorname{TD}(\lambda)$. When applying the idea of Predecessor Features and Expected Traces \cite{et} to Deep RL in the Cartpole task (Figure 2b), bootstrapping shows improved results. 

\section{Discussion}
We examine a method, based on the SR, for assigning credit to preceding states. We show that this method can also be implemented by learning the expected sum of discounted preceding states (or features) directly. This helps learning performance of the value function in both accuracy and speed. This write up is meant to fully examine the idea of Predecessor Features and accordingly tries to learn this representation directly by looking at a fully bootstrapped version in the linear case. This directly relates the Successor to Predecessor Representation. Please note TD-PF is a special instance of van Hassselt's ET($\lambda, \eta$). 
\printbibliography

\end{document}